\documentclass[review]{elsarticle}

\usepackage{lineno,hyperref}
\usepackage{amsfonts}
\usepackage{amsmath}
\usepackage[pdftex]{color}
\usepackage[font=footnotesize,labelfont=bf]{caption}
\usepackage{subfigure}

\modulolinenumbers[5]

\usepackage{xcolor}

\journal{ISPRS Journal of Photogrammetry and Remote Sensing }

\usepackage{xspace}
\makeatletter
\DeclareRobustCommand\onedot{\futurelet\@let@token\@onedot}
\def\@onedot{\ifx\@let@token.\else.\null\fi\xspace}

\def\eg{\emph{e.g}\onedot}

\makeatother

\biboptions{authoryear}

\newcommand{\bn}{\mathbf{n}}
\newcommand{\bc}{\mathbf{c}}

\newcommand{\bP}{\mathbf{P}}

\newcommand{\ba}{\mathbf{a}}

\newcommand{\by}{\mathbf{y}}
\newcommand{\bmf}{\mathbf{f}}
\newcommand{\bp}{\mathbf{p}}

\newcommand{\bbR}{\mathbb{R}}

\begin{document}

\begin{frontmatter}

\title{Deep Learning Guided Building Reconstruction from Satellite Imagery-derived Point Clouds }

\author{Bo Xu$\dagger$$^1$, Xu Zhang$\ddagger$$^1$\footnote{$^1$Equal Contribution.}, Zhixin Li$\dagger$, Matt Leotta$\mathsection$, Shih-Fu Chang$\ddagger$, Jie Shan$\dagger$}
\address{}

\author[mymainaddress]{$\dagger$Purdue University}
\cortext[mycorrespondingauthor]{Corresponding Author: Jie Shan}
\ead{jshan@purdue.edu}
\author[mysecondaryaddress]{$\ddagger$Columbia University}
\author[mythirdaddress]{$\mathsection$Kitware, Inc.}

\address[mymainaddress]{550 W Stadium Ave, West Lafayette, IN, USA}
\address[mysecondaryaddress]{116th St, Broadway, New York, NY, USA}
\address[mythirdaddress]{1712 Route 9, Suite 300, Clifton Park, NY, USA}

\begin{abstract}
3D urban reconstruction of buildings from remotely sensed imagery has drawn significant attention during the past two decades. While aerial imagery and LiDAR provide higher resolution, satellite imagery is cheaper and more efficient to acquire for large scale need. However, the high, orbital altitude of satellite observation brings intrinsic challenges, like unpredictable atmospheric effect, multi view angles, significant radiometric differences due to the necessary multiple views, diverse land covers and urban structures in a scene, small base-height ratio or narrow field of view, all of which may degrade 3D reconstruction quality.  
To address these major challenges, we present a reliable and effective approach for building model reconstruction from the point clouds generated from multi-view satellite images. 
We utilize multiple types of primitive shapes to fit the input point cloud. Specifically, a deep-learning approach is adopted to distinguish the shape of building roofs in complex and yet noisy scenes. For points that belong to the same roof shape, a multi-cue, hierarchical RANSAC approach is proposed for efficient and reliable segmenting and reconstructing the building point cloud.
Experimental results over four selected urban areas (0.34 to 2.04 sq km in size) demonstrate the proposed method can generate detailed roof structures under noisy data environments. 
The average successful rate for building shape recognition is 83.0\%, while the overall completeness and correctness are over 70\% with reference to ground truth created from airborne lidar. As the first effort to address the public need of large scale city model generation, the development is deployed as open source software. 
\end{abstract}


\end{frontmatter}

\section{Introduction}
\label{sec:intro}
\subsection{Motivations}
\label{subsec:motivation}
3D reconstruction of large-scale urban scenes has become an essential task for various applications, such as urban planning, virtual reality, emergency management, and other smart and healthy city related activities. Since reconstructing the 3D models of the urban region requires specific expertise and great human efforts, efficient and automatic reconstruction of building models of large scale scenes has attracted significant attention in recent years~\citep{HAALA2010570,Musialski2013Survey,huang2017towards,Lafarge2016towards}. The extraction of building roofs is confronted with many challenges including complexity of building roofs, data sparsity, occlusion and noise~\citep{Lafarge2015LOD}.

3D model reconstruction generally starts with point cloud. With the current data acquisition techniques as well as the recent improvement in dense matching methods, point clouds from LiDAR data or aerial images are of high precision and density, which helps reconstruct high quality 3D building models. However, in many scenarios, collecting aerial data (LiDAR or imagery) is expensive, time-consuming, less efficient, and sometimes can be risky and impractical. 

Satellite imagery, as an alternative, is much cheaper and easy to access. For satellites like Worldview 3, the spatial resolution can be as high as 0.31m. Using those images for 3D reconstruction is very appealing. Indeed, there already exist several solutions for generating point clouds from multi-view satellite images~\citep{Vricon,P3D}.
However, compared to the point cloud generated by either aerial imagery or LiDAR, the quality of the point cloud from satellite images is often inferior in terms of precision and noise level. 
Moreover, the distribution of the point cloud derived from satellite images tends to be intrinsically different from that of LiDAR data. These makes the building reconstruction from satellite images much more challenging. And it is impractical to directly adopt the existed reconstruction method designed for aerial data to the satellite data.

Under such considerations, this work aims at developing a robust approach to reconstruct building models at a large (e.g., city) scale from point clouds generated by satellite images.

\subsection{Related Works}
\label{subsec:related}
 
\subsubsection{Building Reconstruction from Point Clouds}

The extraction of building roof and their reconstruction strategies mainly converge into three main categories~\citep{Vosselman2010LiDARbook}: model-driven, data-driven, and mix-driven by combining the former two.

Model-driven methods adopt a top-down strategy~\citep{HENN201317Model_driven, Vanegas2012Model_driven,lafarge2011Model_driven}. These kinds of methods need to define a library of roof models beforehand and search typical roof shapes from the library by matching and fitting them to the input point cloud. Therefore, the shapes of the reconstructed model are mostly decided by the way the roof model library is defined. 
Since searching the roof model directly from the point cloud is often time-consuming, the predefined roof model needs to be simple enough but meanwhile adaptive to the real-world complex roofs. \citet{Lafarge2010FromDSM} find the optimal 3D rectangles based on Bayesian decision with a Markov Chain Monte Carlo sampler, where most models are represented as combination of rectangles roofs or gables.
\citet{Vanegas2012Model_driven} use the Manhattan-world to describe the roof structure, where the reconstructed building is grid-like. 
The grid or rectangle like models are oversimplified for real world buildings. The model-driven method suffers when the targeted roof is not in the predefined library, especially for complex urban building roofs. 

On the other hand, data-driven methods adopt a bottom-up strategy, which starts from searching low-level features, such as lines or roof segments~\citep{Verma2006Topology, Elberink2009Topo,Sampath2010Seg}. 
The reconstructed roof structure is then composed by the combination of lower level features. The data-driven approach based on point cloud segmentation is popular when the roof structure is complex or the point density is high. Generally, the roof plane is extracted first, then the ridges and corners are constructed by considering the topology of the plane. A roof topology graph~(RTG) is often used when considering the roof topology. \citet{Verma2006Topology} firstly add labels to RTG to distinguish the type of connections. \citet{Elberink2009Topo} extend it by adding more features like being convex/concave or not, and being horizontal/vertical or not.
\citet{Elberink2009Topo,PERERA2014213} further utilize graph analysis in roof topology analysis. The work~\citep{BoXu2017HRTT} defines a tree structure on the RTG which aims at analyzing the plane-model and model-model relations. 
The data driven method can handle any kind of roofs in theory. However, when decomposing a complex roof, especially a curved roof, it may end up with over-segmentation or under-segmentation, which leads to over-simplified or bulky reconstructed models. 
%

The mix-driven method combines the advantages of both the model-driven and the data-driven approaches. It applies the model-driven approach to generate integral constraints for the normalized structure and then utilizes the data-driven approach to describe various model shapes. 
In fact, many data-driven methods also consider the knowledge of the roof model, such as the model primitives and the roof topology. For instance, \citet{XIONG2015275} assume that the roof primitives consist of planes which belong to the same loop in RTG. 

\subsubsection{Deep Learning for Point Cloud Processing}
Processing point cloud data with deep neural networks has become a hot research topic recently. Typical convolutional neural network~(CNN) structures take highly structured voxelized data as input and used 3D convolution to process the voxel data~\citep{wu20153d}. However, due to the complexity of the 3D convolution, the resolution of the voxel is constrained. 
Multi-view CNN~\citep{su_multi-view_2015} projects the point cloud into multiple 2D images with different view angles and processes multi-view images with multi-brunch 2D CNNs. However, the internal information of the point cloud is often missing due to the necessary projection involved.  
\citet{qi_pointnet:_2016} propose PointNet which directly takes raw point cloud data as input. PointNet and its multi-scale variant PointNet++~\citep{qi_pointnet++:_2017} show strong performance in both 3D point cloud classification and segmentation. 
VoxelNet~\citep{zhou2018voxelnet} combines the voxel and the PointNet. 

Deep neural networks have also been actively applied to remote sensing data. 
\citet{wang_lidar_2018} apply a DNN to object classification in a LiDAR point cloud. 
\cite{zeng2018neural} apply a DNN for 3D reconstruction of residential buildings. 
However, their approach only deals with residential buildings with rather simple structures and needs detailed annotation for the shape and the cross-section of the building. 

The structure of this paper is organized as follows. Section~\ref{sec:overview} is an overview of the proposed reconstruction method. The deep learning based roof shape segmentation method is described in Section~\ref{sec:shape_segmentation}. In Section~\ref{sec:Hierarchical_RANSAC}, we explain how we generate robust roof models based on multi-cue hierarchical RANSAC. Experiments and discussion are provided in Section~\ref{sec:experiment}, followed by the conclusion in Section~\ref{sec:conclusion}.

\section{Proposed Approaches}
\label{sec:overview}

We addresses the urban scene 3D reconstruction problem by using several different types of primitive shapes (such as plane, sphere and cylinder) to fit the point cloud. A deep learning based roof shape segmentation model is proposed to predict the shape of primitives for each point in the point cloud. After that, an iterative RANSAC method is proposed to fit the labeled points with primitives of the predicted shape. 

To deal with the high noise level in the satellite image-derived point cloud, the deep learning based roof shape segmentation is directly learned from satellite image-generated point clouds to ensure the segmentation quality.
To effectively collect the training data, we further propose a data augmentation method which can easily synthesize realistic complex building roofs with different shapes. We further propose a multi-cue hierarchical RANSAC to fit proper primitives to the point cloud. The proposed RANSAC method incorporates shape, surface normal, and color information from multiple scales and shows high accuracy and efficiency in dealing with the noisy point cloud.

\begin{figure}[ht]
    \centering
    \includegraphics[width=1\textwidth]{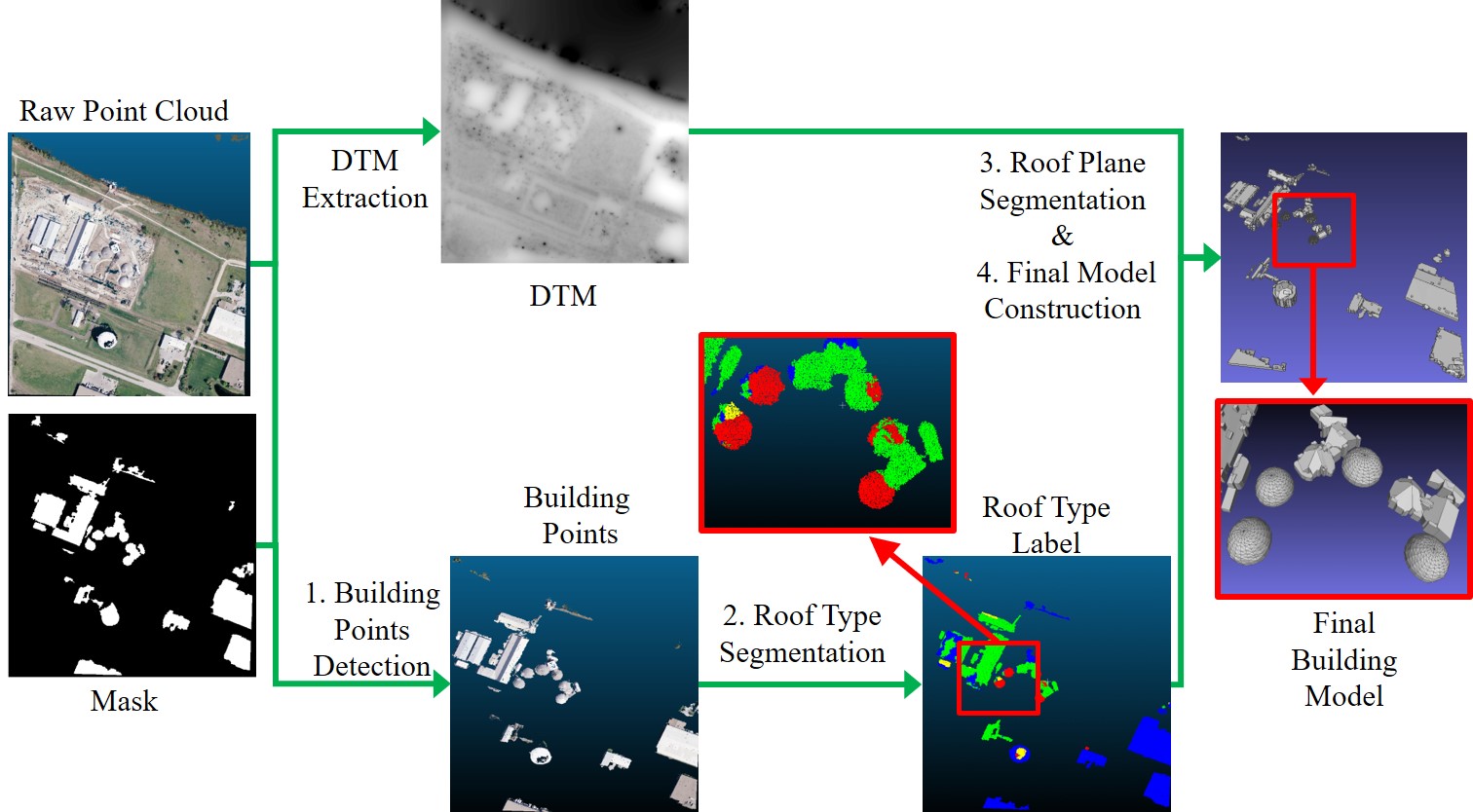}
    \caption{Overall workflow of the building reconstruction strategy}
    \label{fig:overview}
\end{figure}

As shown in Fig.~\ref{fig:overview}, the input of our approach consists of two parts: 
1) A point cloud generated through stereo matching of high resolution satellite images. The point cloud is a set of points $\bP_{all} = \{\bp_i\}$, $i  = \{1,\ldots,N\}$, where $\bp_i \in \bbR^6$ is a single point in the point cloud with six dimensions, i.e., the geometric coordinate (x, y, z) and the RGB color. 
2) An automatically generated building mask, which is an ortho-rectified binary raster image. Each pixel in the mask indicates if the position belongs to building (1) or not (0). Note that the automatically generated building mask may contain error. The correspondence between a point in the point cloud and its position in an image is given by the RPC (Rational Polynomial Coefficients). 

The goal of 3D building reconstruction is to find a set of primitive shapes (such as: plane, sphere and cylinder) to represent the 3D shape of the building in the point cloud.
We first generate the Digital Terrain Model~(DTM) by terrain filtering upon the point cloud by the Cloth Simulation Filtering~(CSF) method~\citep{zhang2016dtm}. The DTM is also a ortho-rectified raster image in which each pixel indicates the height of the ground at that position.

To reconstruct the building model, we first detect the building points in the point cloud by selecting points laid in the building mask. All the building points are divided into different clusters via Euclidean cluster extraction~\citep{cgal}(step 1). 
We then recognize different shapes in the point cloud via a deep neural network. The network takes a point cluster as input and outputs the shape type for each point(step 2).
For points that have the same shape type within each point cluster, a hierarchical RANSAC method is proposed to extract the primitive shape with location, size and orientation(step 3) to fit the points. 
The boundary of the primitive shape is determined by using the roof topology~\citep{BoXu2017HRTT,sampath2007building}~(step 4). 

The main contributions of this paper are:
\begin{enumerate}
    \item Proposing an end-to-end approach to reconstruct the 3D building model from satellite image-generated point clouds with multiple types of primitive shapes.
    \item Applying a deep learning based method for roof shape segmentation and proposing a data augmentation method to effectively collect the building roofs with different shapes. 
    \item Proposing multi-cue hierarchical RANSAC to extract shapes from noisy point cloud data. 
    \item Demonstrating satisfactory 3D reconstruction results from the proposed pipeline with several large scale urban areas.
\end{enumerate}

\section{Roof Shape Segmentation using Deep Learning }
\label{sec:shape_segmentation}
Building roofs can be very complex in the real world and may consist of different shapes of surfaces (e..g., planar, cylindrical and spherical). Most of the previous works use multiple planar surfaces to approximate the curved surfaces~\citep{cao20173d,huang2017towards}. However, this leads to a fractured results consisting of many small and narrow planar surfaces. It is natural and more meaningful to decompose the complex roof into a few basic primitive shapes such as plane, cylinder and sphere~\citep{sharma_csgnet:_2017}.
However, due to the high level of the structured noise as well as the location errors in the satellite image-generated point cloud, directly decomposing the point cloud using geometric constraints is very challenging. To resolve this problem, we propose to train a deep learning-based roof shape segmentation network with the satellite image-generated point clouds directly. Given the point cloud as input, the segmentation network assigns one shape type label to each point in the point cloud. 

\subsection{Roof Shape Segmentation}
\label{subsec:roof_shape_segmentation}
The roof shape segmentation model aims at learning a function, $\bmf(\cdot)$, which takes a point cloud  $\{\bp_i\}_{i = 1,\ldots,n}, \bp_i \in \bbR^d$ as input and outputs a set of one-hot vectors $\{\by_i\}_{i = 1,\ldots,n}$, $\by_i \in \{0,1\}^L$, where $\parallel \by_i \parallel_2 = 1$ is the shape indicator for point $\bp_i$ in the input point cloud, and $L$ is the number of types of the shape.
\begin{equation}
    \bmf(\{\bp_1,\ldots,\bp_n\}) := \{\by_1,\ldots,\by_n\}
\end{equation}
The point cloud has two important properties. 1) It is an unordered set of points, which means no matter how the input order of the point changes, the point cloud is still the same point cloud. 2) Each point in the point cloud is not isolated. The relative location of the neighboring points defines the shape. The roof shape segmentation model should be able to consider these two properties. We find that PointNet~(\cite{qi_pointnet:_2016}) fulfills the requirement to be the segmentation model. 

\subsection{Overview of PointNet}
\label{subsec:review_pointnet}
 \begin{figure}[ht]
    \centering
    \includegraphics[width=0.9\textwidth]{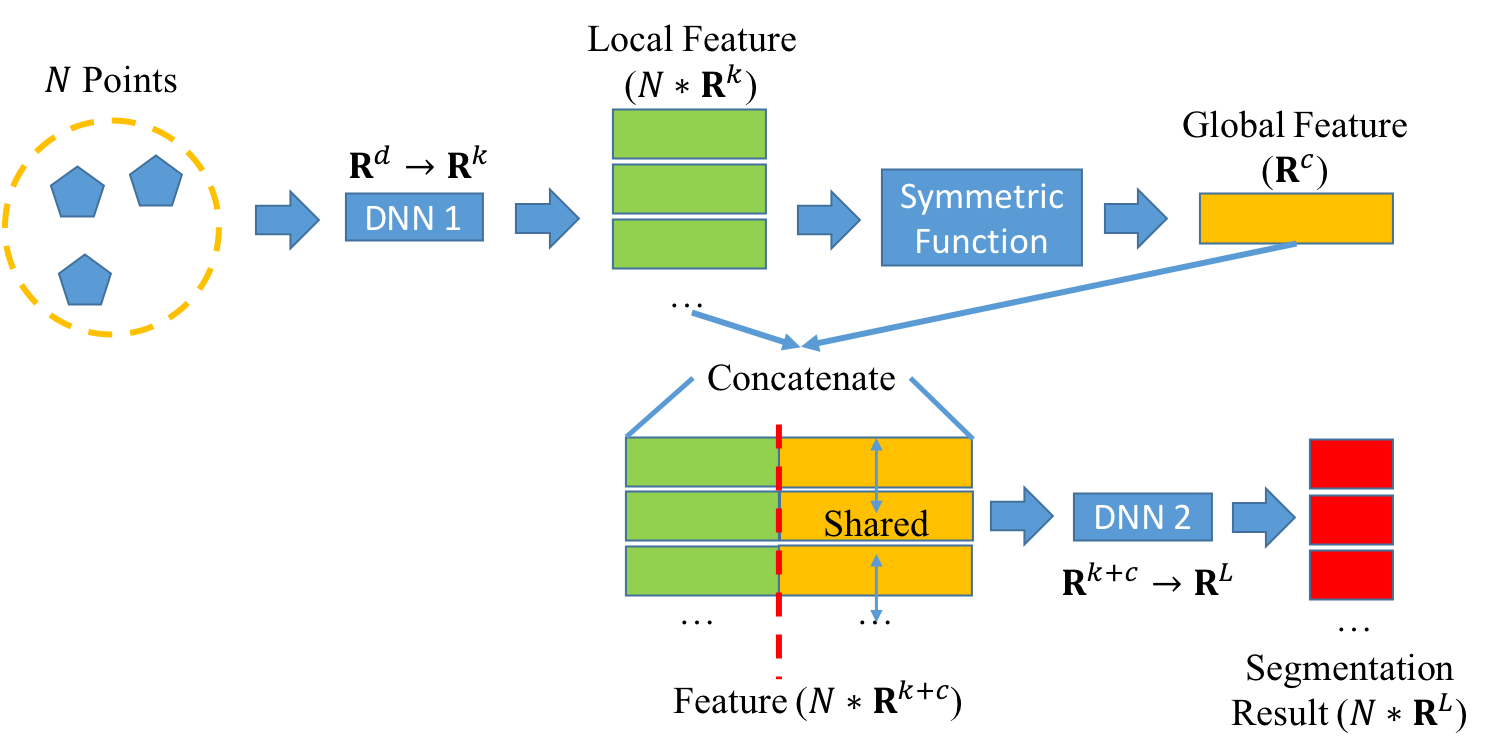}
    \caption{The basic pipeline of the PointNet~\citep{qi_pointnet:_2016}}
    \label{fig:shape_segmentation}
\end{figure}

\citet{qi_pointnet:_2016} proposed the powerful and effective PointNet model to solve the point cloud segmentation problems. The pipeline of the segmentation model is shown in Fig.\ref{fig:shape_segmentation}. 
Given a point cloud of $N$ points, each point passes through the first neural network which contains a few transform layers and fully connected layers to get one $k$ dimensional feature for each point.  
A symmetric function (\eg element-wise max pooling) is applied to the features of the $N$ input points. The output of the symmetric function is a $c$ dimensional global feature, which is the representation of the whole input point cloud. Due to the symmetric function, the global feature is invariant to the input order of the points. In other words, no matter how the order of the input changes, the global feature remains the same. 

The global feature is then concatenated with each local feature. Thus the concatenated feature contains both the local and the global information. The concatenated feature passes through the second neural network, which is a multi-layer perceptron. The network assigns one shape label to each point as the final segmentation result.

\subsection{Synthesizing Realistic Building Roofs}
\label{subsec:synthesize_roof}
Training the roof shape segmentation model requires hundreds of point clouds with detailed shape labels on each point. Collecting such a dataset is impractical. 

One possible solution is to sample points from some standard shapes (such as plane, cylinder and sphere) and use those points as training sample. However, since the satellite image-generated point cloud contains high level of structured noise, the roof shape is often considerably different from the standard shapes. The model trained with standard shapes may not generalize well to the point cloud.

Collecting training data with labels from point clouds is important to guarantee the accuracy of the segmentation model. Unfortunately, collecting point clouds with different shapes is not an easy task, since most of the residential buildings have flat or sloped roofs. In order to effectively collect roofs with different shapes, we propose to synthesize other shapes of roofs, especially cylindrical and spherical roofs, from flat roofs. 

\begin{figure}[ht]
\begin{center}
\subfigure[RGB image of the roof]{
	\includegraphics[width=0.20\linewidth]{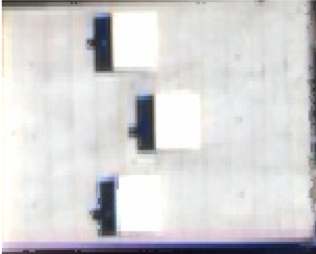}
	\label{fig:Roof_RGB}
	} 
\hspace{2.2em}
\subfigure[Side view of the cropped points]{
	\includegraphics[width=0.26\linewidth]{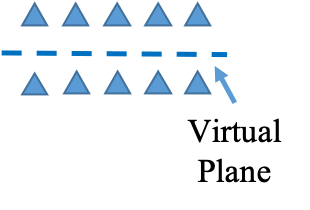}
	\label{fig:cropped_points}
	}
\hspace{2.0em}
\subfigure[Side view of the synthesized points]{
	\includegraphics[width=0.28\linewidth]{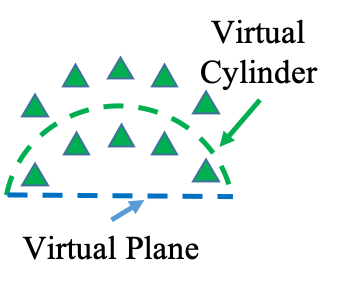}
	\label{fig:bended_points}
} 
\\
\subfigure[Transformation of the point cloud in 3D.]{
	\includegraphics[width=0.9\linewidth]{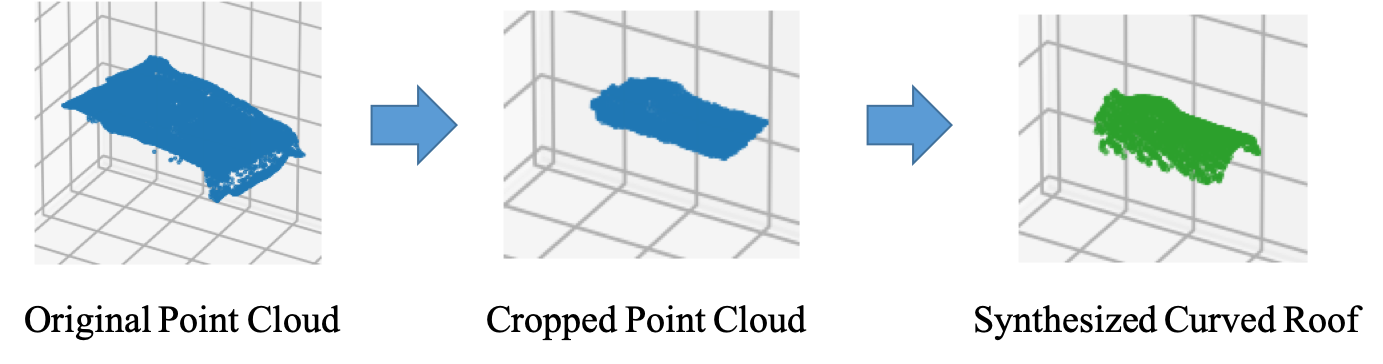}
	\label{fig:pipeline}
}	
\vspace{-1.0em}
\caption{Synthesize realistic cylindrical roof from flat roof.}
\label{fig:synthesized_roof}
\vspace{-1.0em}
\end{center}
\end{figure}

To synthesize a cylindrical roof(Figure~\ref{fig:pipeline}), given a flat roof, we first crop points within a randomly selected rectangular region that is parallel to the ground. We assume the cropped points are sampled from a flat rectangle roof of height $h_0$, which is the average height of the cropped point. The actual height of each point with respect to the flat plane of height $h_0$ reflects the noise of the point cloud.
We synthesize a cylindrical roof by bending the flat roof. Firstly, a cylinder that is also parallel to the ground with random radius is generated by restricting the rectangle as a cross section of the cylinder.  Assuming the equation of the cylinder is $z = g(x,y)$, we move the original point up for distance of the height between the cross section and the cylinder. 
Mathematically, each point $(x,y,z)$ in the cropped point cloud is moved to $(x,y,z')$, where
\begin{equation}
    z' = z - h_0 + g(x,y)
\end{equation}
Therefore, the new point cloud has a cylindrical shape which preserves the original noise of the flat roof.

For other shapes of roof, the synthesis process is similar. For example, for spherical roofs, instead of cropping a rectangular region, we crop a circular region and then bend the plane to a sphere. 

We combine the synthesized point clouds with different shapes to make complex roofs and use them to train our roof shape segmentation model. The loss function is the cross-entropy loss. 

Experiments in Sec.~\ref{sec:experiment} show that the synthesized building roof very well reflects the distribution of the point cloud generated by satellite images. A model trained with the synthesized building roof point clouds achieves much better performance than the model trained with the point clouds sampled from standard shapes. 

After identifying the roof shape in the point cloud, we yet need to determine the parameters of the primitives. In our practice, we found that spherical and cylindrical roofs can be modeled well with the conventional iterative RANSAC. However, the combination and the intersection of the planar (flat and sloped) roofs are more challenging to deal with. As such, we propose a multi-cue hierarchical RANSAC technique based on color, shape, and normal to determine their parameters from the shape-labeled point cloud.
%

\section{Planar Roof Segmentation with Augmented RANSAC} 
\label{sec:Hierarchical_RANSAC}
Given a point cloud and the shape label of the point cloud, we augment the classical RANSAC method by introducing  a multi-cue and hierarchical strategy to estimate the parameters for the planar roof primitives that best fit the point cloud. 

\subsection {Overview of RANSAC}
\label{subsec:RANSAC}
Given a point cloud $\{\bp_i\}_{i = 1,\ldots,n}, \bp_i \in \bbR^d$ to fit with a specific model, the RANSAC algorithm recursively selects a minimum set of random points to solve a model with parameter $\hat{\ba}$. The solved model is then tested through all the points in the point cloud to see how well the model fits the point cloud. The fitting score, indicating how good the fitting is, is defined as:
\begin{equation}
   S = \sum_{i=1}^n W(\bp_i, \hat{\ba}) I(\bp_i, \hat{\ba})
\end{equation}
where $I(\bp_i, \hat{\ba})$ is an indicator function to see whether $\bp_i$ is an inlier of $\hat{\ba}$ or not, $W(\bp_i, \hat{\ba})$ is a weight function showing how well the point fits the model. In conventional RANSAC, $W(\bp_i, \hat{\ba})=1$. 
The algorithm runs multiple times to find the best hypothesis with the highest score. After that, all the inliers of the best hypothesis are used to estimate a new model as the final model. After removing the inliers from the point cloud, the RANSAC can be iteratively applied to the remaining points to get the new fitting models until the remaining points are fewer than a threshold. 

Since point clouds generated from satellite imagery may contain high noise, directly applying the conventional RANSAC algorithm to the point cloud may lead to over-segmentation. To improve the robustness of RANSAC, we introduce multi-cue hierarchical RANSAC which incorporates color, shape, and normal information in a coarse-to-fine manner.

\subsection {Multi-Cue RANSAC}
\label{subsec:weightedRansac}
In ~(\cite{BoXU2016wRANSAC}), the point-to-plane distance, and the angle between the point normal and the model normal are gathered as a joint weight to evaluate the contribution of a point $\bp$ to a hypothesis model $\hat{\ba}$.  
The weights for the distance and the angle between the normal vectors are given below
\begin{equation}
\begin{aligned}
W_{dis}(\bp, \hat{\ba}) & =  e^{- d(\bp, \hat{\ba})^2/\sigma_{dis}^2}, \\
W_{nv}(\bp, \hat{\ba})  & =  e^{-\parallel\bn(\bp) - \bn(\hat{\ba}, \bp)\parallel^2/\sigma_{nv}^2} 
\end{aligned}
\label{eq:weights_disandnv}
\end{equation}
where $d(\bp, \hat{\ba})$ is the Euclidean distance between the $\bp$ and $\hat{\ba}$, $\bn(\bp)$ is the normal vector of $\bp$ estimated from its nearby points and $\bn(\hat{\ba}, \bp)$ is the normal vector of the model $\hat{\ba}$ at the point that is closest to $\bp$. $\sigma_{dis}$ and $\sigma_{nv}$ are two trade-off parameters. 

For satellite image-based point clouds, since points within the same plane tend to have similar material and reflectance, the color similarity between the hypothesis model and the point should also be taken into consideration. The weight for the color is defined as  
\begin{equation}
W_{rgb}(\bp, \hat{\ba})=e^{-\parallel\bc(\bp) - \bc(\hat{\ba})\parallel^2/\sigma_{rgb}^2}
\label{eq:weights_rgb}
\end{equation}
where $\bc(\bp)$ is the color vector (R,G,B) of $\bp$, $\bc(\hat{\ba})$ is the color vector of the model $\hat{\ba}$ which is defined as the average RGB value of its seed points (points used to estimate the model). $\sigma_{rgb}$ is the trade-off constant for color.

The final weight of a single point with respect to a model is defined as the multiplication of the above three weights:
\begin{equation}
W(\bp, \hat{\ba}) = W_{dis}(\bp, \hat{\ba})W_{nv}(\bp, \hat{\ba})W_{rgb}(\bp, \hat{\ba})
\label{eq:weights_pt}
\end{equation}

\subsection {Multi-Cue Hierarchical RANSAC}
\label{subsec:Hierarchical_RANSAC}
\begin{figure}[ht]
\centering
\includegraphics[width=0.9\textwidth]{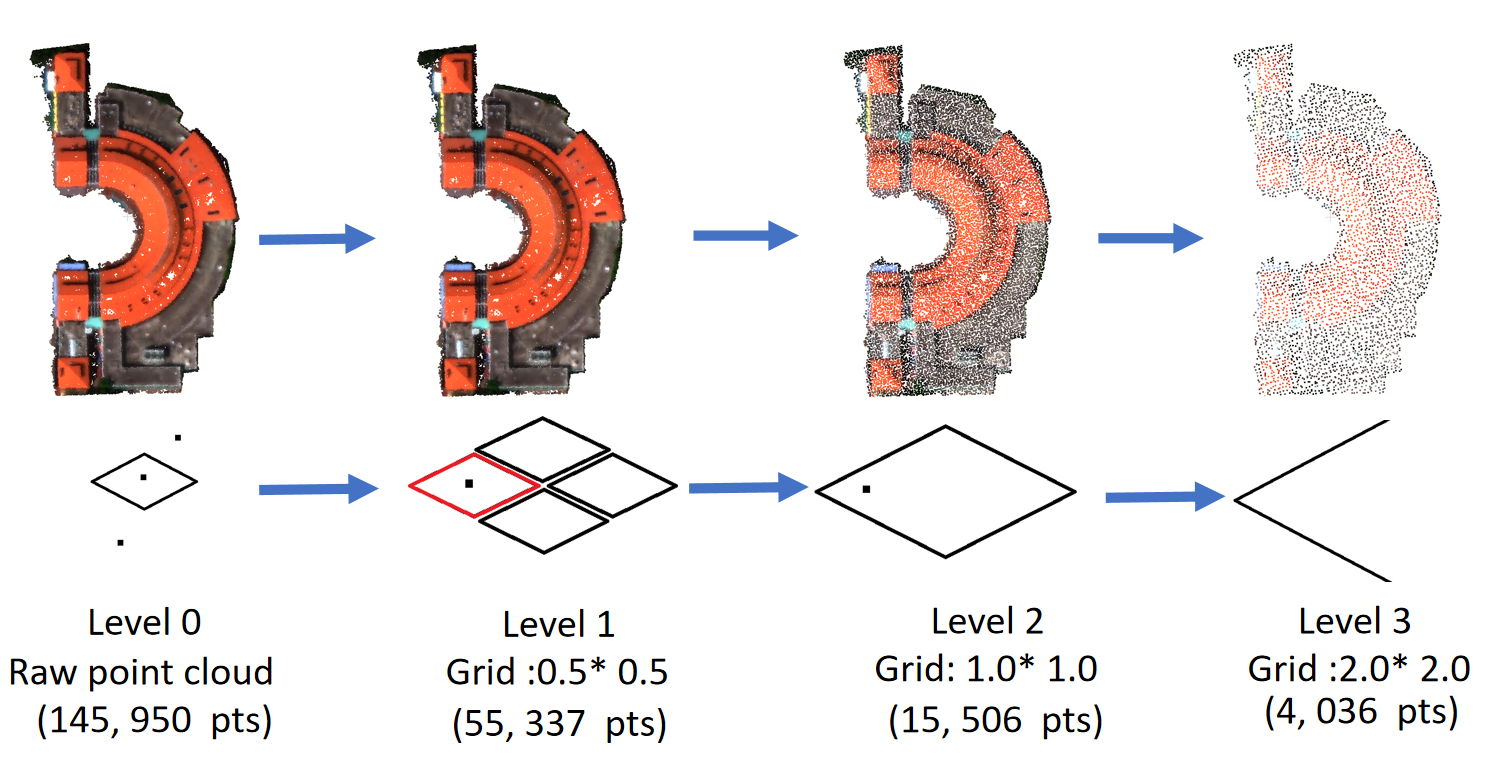}
\caption{The structure of point cloud pyramid for hierarchical RANSAC}
\label{fig:H_RANSAC}
\end{figure}

To further improve the stability of the RANSAC algorithm, we propose a hierarchical structure for the RANSAC method. It down-samples the input point cloud step-by-step to form a pyramid structure, as shown in~\ref{fig:H_RANSAC} and extracts the model parameters from coarse to fine. 
The raw point cloud is regarded as the first (finest) level of the point cloud pyramid. After smoothing and the median filtering, a 2*2 down-pooling filter~(we used 0.5m*0.5m for the raw point cloud) is applied to the point cloud. Only the point with the median height in each grid is retained. This helps ensure the point density and mitigate the influence of noise. The filtered point cloud is regarded as the next level of the pyramid. This procedure is repeated until a predefined maximum number of levels (usually 3) is met. 

Once the pyramid is constructed, we use the multi-cue RANSAC mentioned above to segment the point cloud from top (coarse) to the bottom (finer) of the pyramid. The ratio of the segmented points to all the points, the minimum number of the points in one single roof, and the mean square error (MSE) of the fitted plane are used as thresholds. 
We iteratively run the algorithm to extract roof primitives until any of the above thresholds is met. Strict thresholds are used at higher level for only detecting robust and large roof primitives. Once the threshold is met in one level, we move to the next lower level, where only the points that are not considered by the previous model will be taken into account.  
Specifically, one point in the higher level may correspond to 1~4 points in the current level. For a higher level point that is fitted to a model, if the distance to the model of any of its corresponding points in the current level is less than a threshold, the corresponding point is considered as being fitted  by the model and will not participate in the segmentation procedure in the current level. 

The major advantage of using such a hierarchical strategy is that the spurious details can be omitted in higher pyramid levels, thus large primitives can be extracted first with high confidence. Such multiple scale/resolution strategy can also improve the algorithm efficiency since the amount of points is much smaller in high pyramid levels and the details are only processed in the remaining data set. In practise, the hierarchical RANSAC is only applied to flat and sloped roofs. For the curved roofs, the traditional iterative RANSAC seems to work well.  

After that, We use the alpha-shape hull to trace the boundary~\citep{sampath2007building} and the roof topology graph~\citep{BoXu2017HRTT} for the intersection of different shapes. Given the segmented roof surfaces and local DTM, building facades can be created by draping roof edges to the ground. Finally, building models are reconstructed by the assembly of top roof, facades and ground.

\section{Experiments and Discussion} 
\label{sec:experiment}
This section will analyze and evaluate the performance of the proposed method. Various assessment metrics are introduced and various regions with complex roof shapes are utilized to test the overall performance of the system. Our implementation is publicly available as part of the Kitware Danesfield repository\footnote{\url{https://github.com/Kitware/Danesfield}}~\citep{leotta2019urban}.
 
\subsection{Data and metrics}
\label{subsec:data_metric}

\begin{figure}[ht]
\begin{center}
\subfigure[AOI 1]{
	\includegraphics[height=0.4\linewidth]{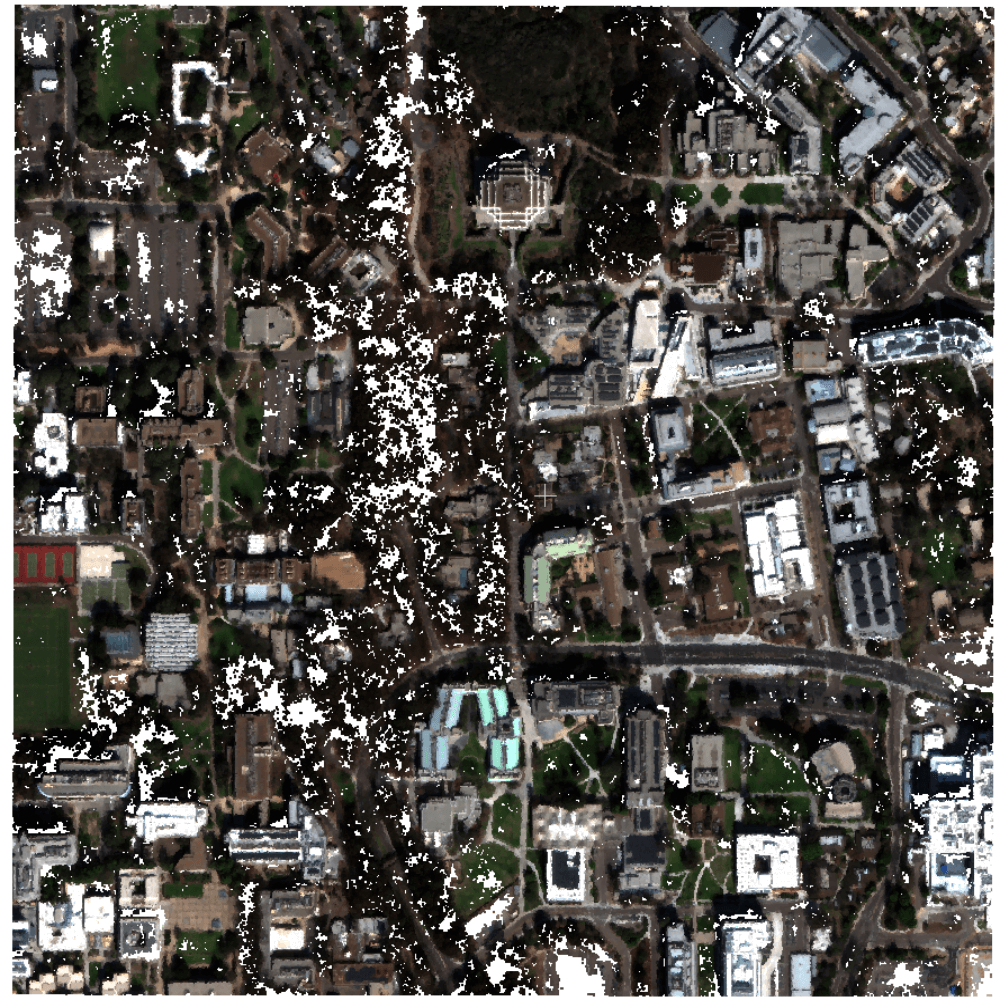}
    \label{fig:Data_AOI1}
	}
\subfigure[AOI 2]{
	\includegraphics[height=0.4\linewidth]{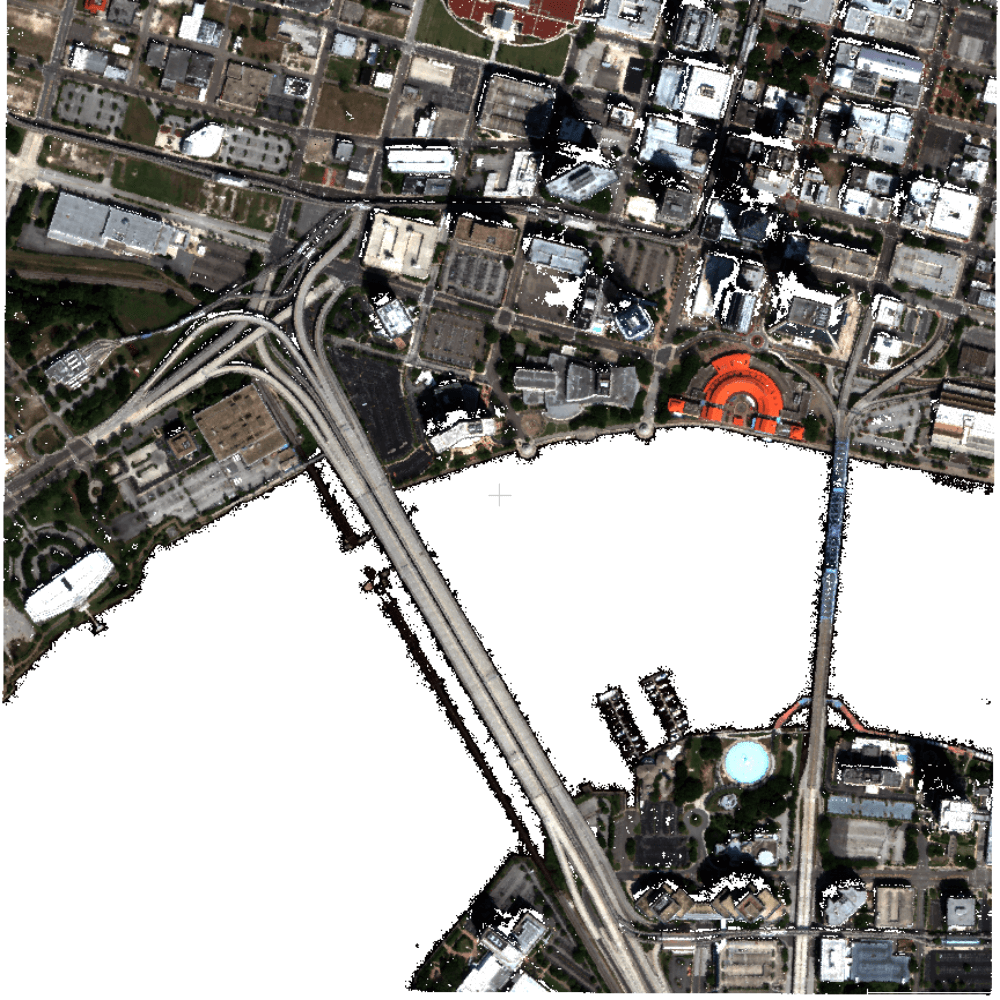}
	\label{fig:Data_AOI2}
	}
\subfigure[AOI 3]{
	\includegraphics[height=0.4\linewidth]{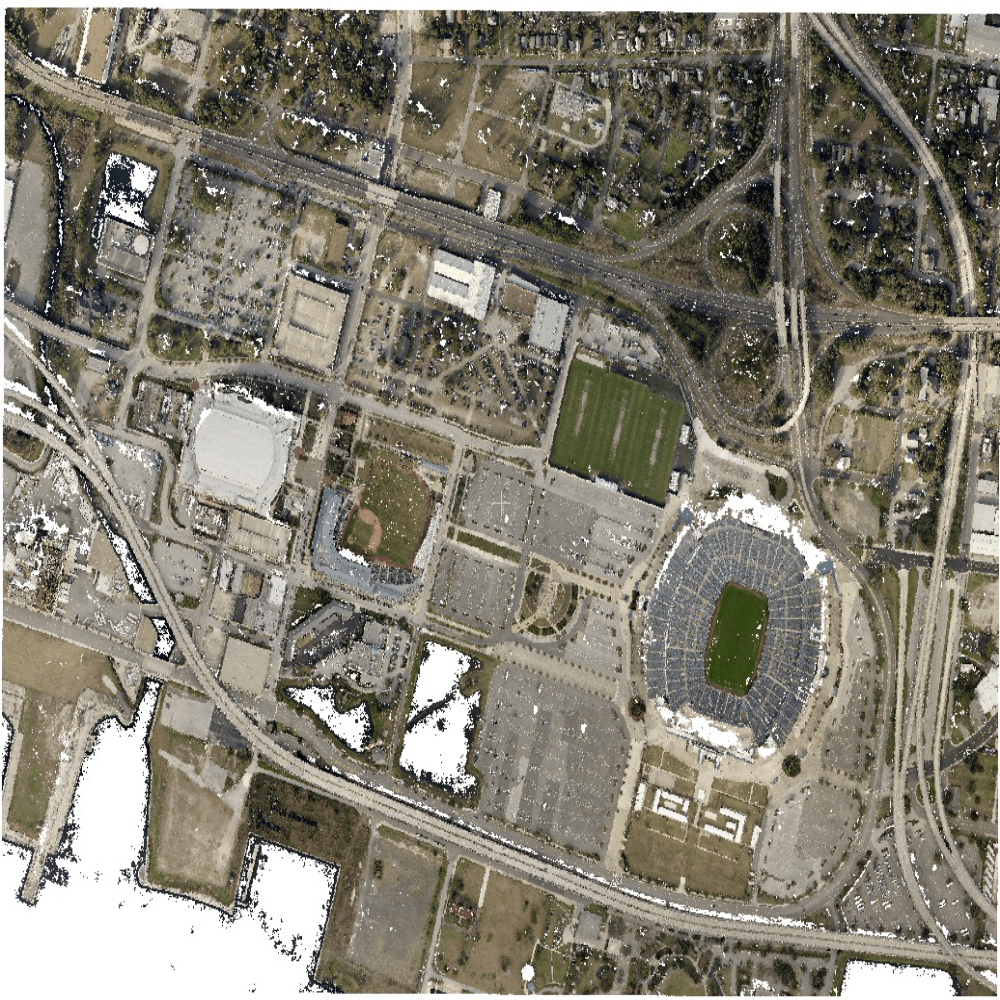}
	\label{fig:Data_AOI3}
	}
\subfigure[AOI 4]{
	\includegraphics[height=0.4\linewidth]{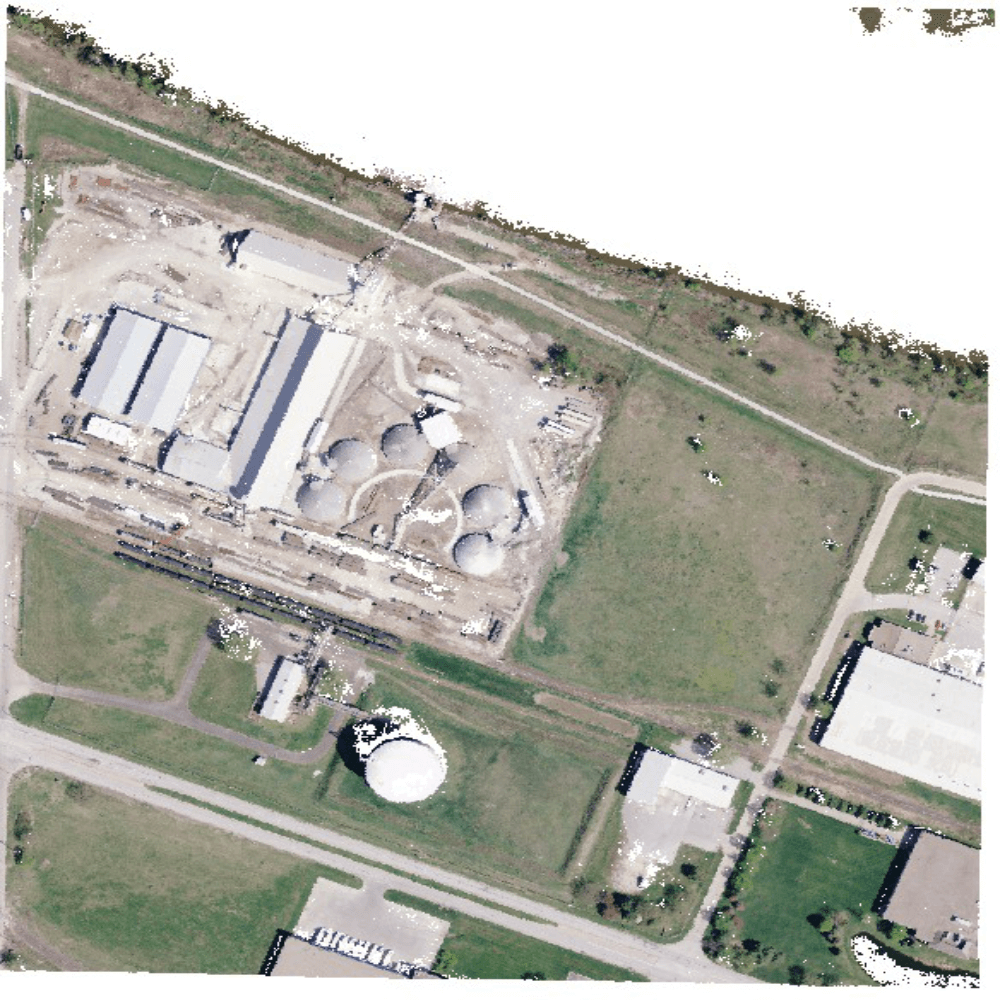}
	\label{fig:Data_AOI4}
	}	
\vspace{-1.0em}
\caption{Raw P3D point cloud for all AOIs. The white part shows the void area (no point).}
\label{fig:AOIs}
\vspace{-1.0em}
\end{center}
\end{figure}

The initial input of the proposed methods are point clouds derived from public available multiple view satellite images~\citep{brown2018core3ddata}. We used the P3D point clouds from the Raytheon company~\citep{P3D}.  As shown in Fig.~\ref{fig:AOIs}, four Areas-of-Interests(AOIs) from different cities in the U.S. are selected. Each point cloud was derived through bundle adjustment and image matching of 15 to 30 WorldView-3 satellite images. 
AOI 1 is selected from the campus of the University of California, San Diego (UCSD), California. The region contains large ratio of vegetation to man-made structures. It is designed to test building extraction and reconstruction algorithms under the occlusion of vegetation.  
AOI 2 is located in the city of Jacksonville, Florida and contains complex bridges and skyscrapers. It is used to test the performance of the reconstruction algorithm in the urban region. 
%
AOI 3 is the TIAA Bank Field in Jacksonville, Florida, which contains a complex outdoor stadium. It tests if the algorithm can deal with complex building shapes.
AOI 4 is Watco Omaha Terminal in Omaha, Nebraska, which contains a few half-sphere shaped warehouses. We use it to test how the reconstruction algorithm handles the spherical roofs.
The statistics of the four AOIs are provided in Table~\ref{tab:Description_data}. The average point densities are between 4.5 to 10.5 points per square meter. All the building masks come from the building segmentation method in \citet{leotta2019urban}.

\begin{table}[ht]
    \small
    \centering
    \begin{tabular}{c c c c c c}
    \hline
    AOI & Location & Area(km*km) & Pts & ${\Delta}$ Z(m) & Pts/$m^2$\\ \hline
    1 & UCSD & 0.99*0.97 & 5,769,279 & 76.04 & 6.01\\ 
    2 & Jacksonville Downtown& 1.41*1.45 & 9,740,605 & 251.77 & 4.76\\
    3 & TIAA Bank Field & 1.47*1.12 & 11,065,390 & 132.07 & 6.75\\ 
    4 & Watco Omaha Terminal & 0.55*0.62 & 2,372,453 & 45.99 & 6.95\\ 
    \hline
    \end{tabular}
    \caption{Detailed information for the four selected regions}
    \label{tab:Description_data}
\end{table}

In order to evaluate the performance of the reconstruction results, independently manually labeled building masks and the Digital Surface Model (DSM) derived from Aerial LiDAR data by \citet{brown2018core3ddata} are provided as reference for AOI 1 and AOI 2. For AOI 3 and AOI 4, we only perform qualitative evaluation.

\subsection{Pre-Processing} 
\label{subsec:pre_pro}
The quality of satellite point cloud is not comparable to the ones from airborne LiDAR or aerial images. The major difficulties exist in the following aspects: low height precision, uneven point density with voids, spurious shadow points. We apply two pre-processing techniques to deal with these issues.

\noindent\textbf{Point Cloud Smoothing} 
The major problem for the satellite image-generated point clouds is the high level structured noise. The RMS of the points within supposed roof plane can be as large as 0.5m. This can greatly influence the precision of plane fitting. We apply the moving least squares algorithm in PCL~\citep{Alexa2004MLS} and median filtering to deal with this.

\noindent\textbf{Holes Filling} 
Also, the point density of stereo matching points is uneven. There are considerable number of ``holes''~(void area) in the point cloud due to the failure of the stereo matching in shadow and non-texture (\eg water and glass surfaces) regions, which introduce challenges for region growing and connectivity checking algorithms and lead to over-segmented sections and holes in the final models.

We first build triangular meshes using the smoothed building points. If any of the triangle mesh of the building is larger than a threshold, we fill the mesh with points of a fixed grid.  



\subsection{Results and Evaluation}
\label{subsec:evaluation}
\subsubsection{Roof Shape Segmentation}
To evaluate the performance of the proposed roof shape segmentation algorithm, we manually annotate the roof shape label for all the buildings in the four aforementioned AOIs. Four shapes of the roofs, including flat (blue), sloped (orange), cylindrical (green) and spherical (red) roofs are considered. 
We generate two different sets of the training data, 1) randomly sample points from the standard shape with different parameters and add Gaussian noise on top of the points (Standard shape); 2) manually select flat roofs and sloped roofs from the point cloud and synthesize cylindrical or spherical roofs using the proposed method.  We use around 300 roofs for each shape type. For the latter training dataset, the selected flat and sloped roofs are not overlapped with the four test AOIs. 
We used ADAM optimizer~(\cite{kingma2014adam}) with a learning rate of 0.001. The learning rate is reduced to 0.7 of the previous value every 20,000 steps. The batch size is 32 and the network is learned for 100 epochs. Rotation, scaling and translation are used for data augmentation. To make a complex roof, 1-3 simple roofs are randomly selected and combined. PointNet++~(\cite{qi_pointnet++:_2017}) is chosen as the based model.

\begin{figure}
\begin{center}
\subfigure[AOI1]{
	\includegraphics[width=0.95\linewidth]{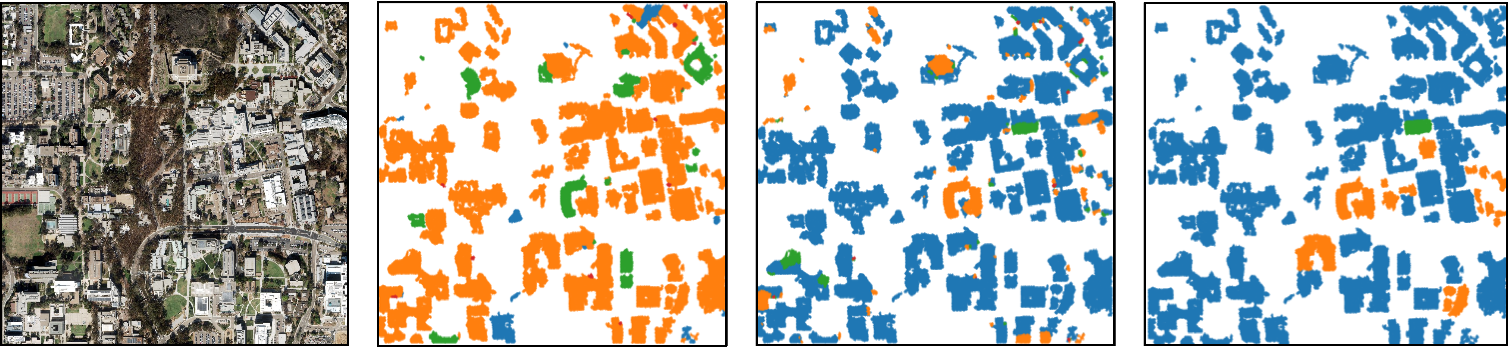}
	\label{fig:Roof_Type_Segmentation_AOI1}
	} \\
\subfigure[AOI2]{
	\includegraphics[width=0.95\linewidth]{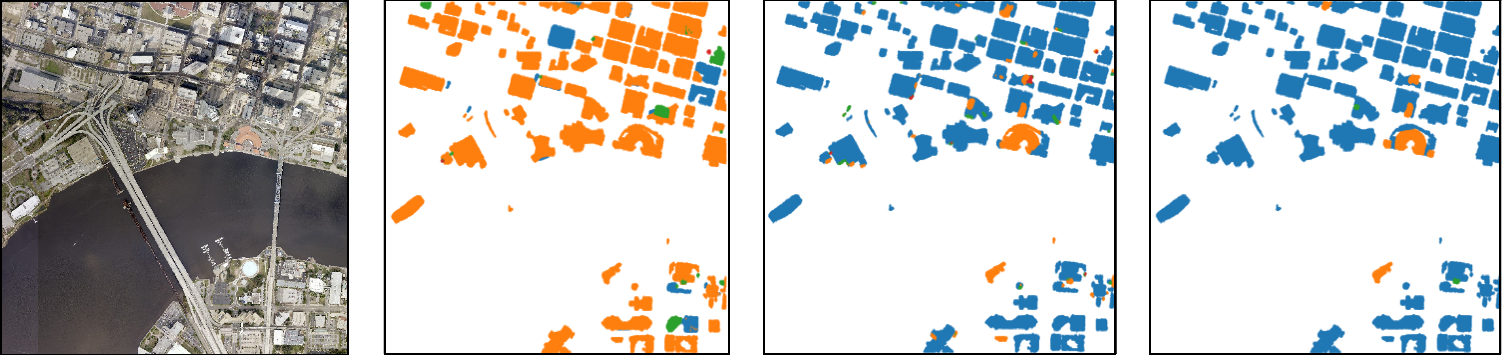}
	\label{fig:Roof_Type_Segmentation_AOI2}
	} \\
\subfigure[AOI3]{
	\includegraphics[width=0.95\linewidth]{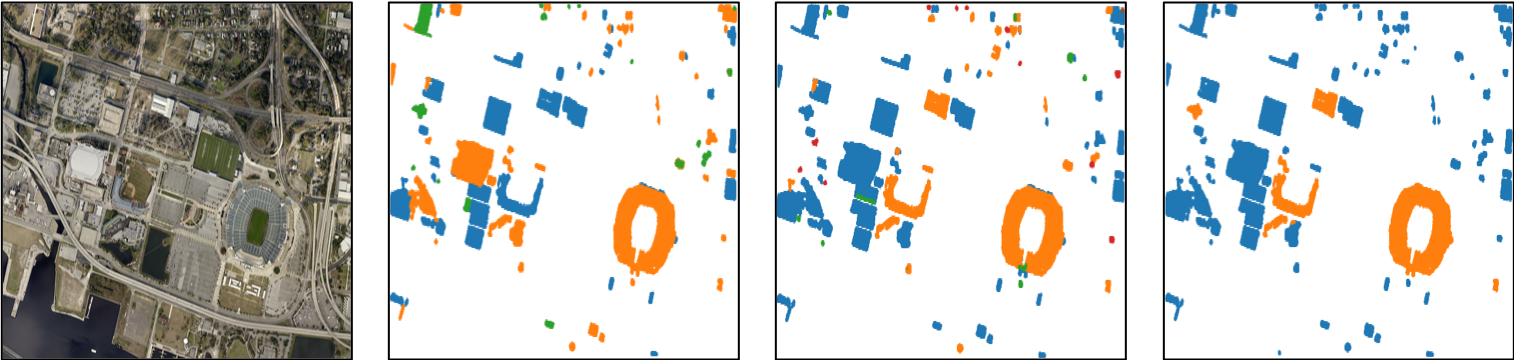}
	\label{fig:Roof_Type_Segmentation_AOI3}
    } \\ 
\subfigure[AOI4]{
	\includegraphics[width=0.95\linewidth]{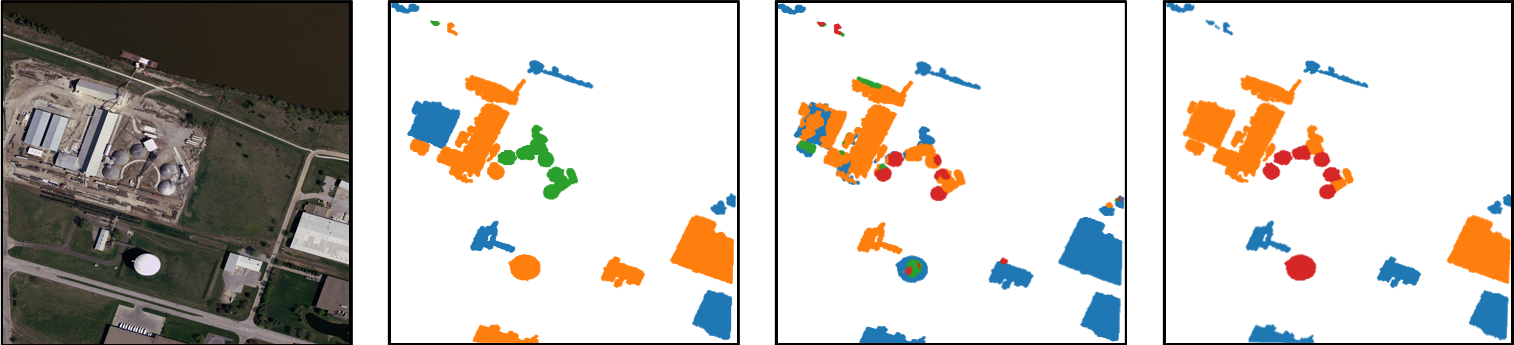}
	\label{fig:Roof_Type_Segmentation_AOI4}
    } \\	
\vspace{-1.0em}
\caption{Qualitative roof shape segmentation results (Different colors represent different roof shape: flat-blue, sloped-orange, cylindrical-green and spherical-red). }
\label{fig:Roof_Type_Segmentation}
\vspace{-1.0em}
\end{center}
\end{figure}

During the test phase, given a point cloud for the whole AOIs, we first run cluster extraction method in PCL~(\cite{Alexa2004MLS}) to separate isolated building point clouds into different clusters based on the Euclidean distance. Each cluster is sent to the segmentation model to assign a shape label to each point. The predicted shape label is compared to the manually annotated label and the prediction accuracy for each AOI is reported in Table~\ref{tab:roof_shape_segmentation_quantitative}. 
We visualize the results in Fig.~\ref{fig:Roof_Type_Segmentation}. The figures from left to right are ortho-rectified RGB image, result predicted by the model learned with standard shape, result predicted by the model learned with our synthesized realistic roofs, and the manually labelled ground-truth. 

The segmentation model trained with the standard shape has inferior performance. From Fig.~\ref{fig:Roof_Type_Segmentation}, we see that the network makes a lot of mistakes by predicting the flat roof as the sloped roof. The reason is that the shape of the point cloud generated from satellite images is not matched well with the standard shape. There may exist attached structures on top of the flat roof and the boundary of the flat roof may be bumpy. Those will mislead the network to recognize the flat roof as sloped roof. 
The model trained with the real roof and our synthesized curved roof has better performance, since it directly learned from the satellite image-generated point cloud. 
With the segmentation result, we fit primitives to corresponding predicted points with our multi-cue hierarchical RANSAC. 

\begin{table}[ht]
   \begin{center}
      \setlength{\tabcolsep}{4.9pt}
      \begin{tabular}{c | c c c c | c}
       \hline
	 Method        & AOI1 & AOI2 & AOI3 & AOI4 &  Ave \\ 
       \hline
	   Standard & 10.5 & 13.1 & \textbf{62.8} & 61.2 & 36.9 \\
       \hline
	   Ours     & \textbf{89.7} & \textbf{91.6} & 57.8 & \textbf{93.0} & \textbf{83.0}\\
      \hline
   \end{tabular}
\end{center}
\caption{Roof shape segmentation accuracy of DNN models trained with different data  (unit: \%).}
\label{tab:roof_shape_segmentation_quantitative}
\end{table}

\begin{figure}[ht]
\begin{center}
\subfigure[AOI 1]{
	\includegraphics[height=0.35\linewidth]{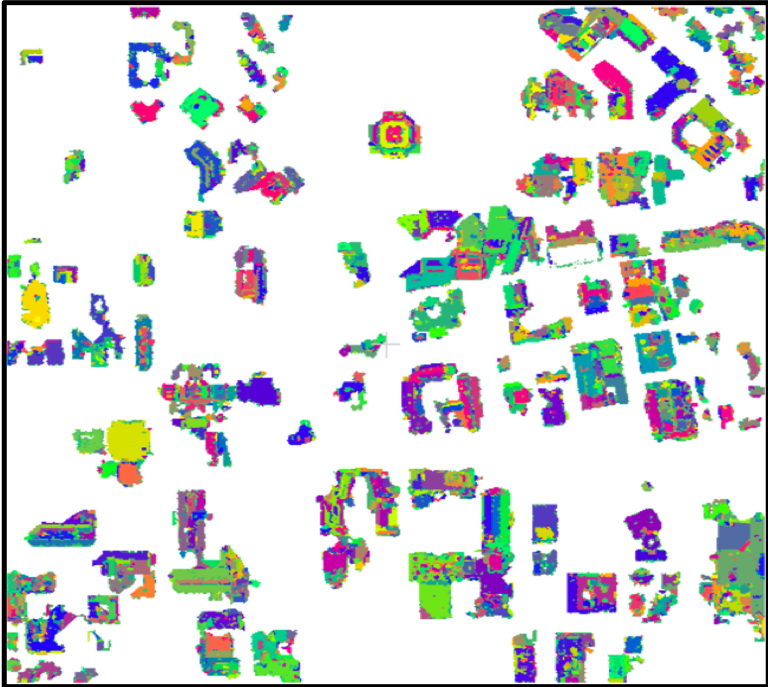}
	\label{fig:segAOI1}
	}
\hspace{2em}
\subfigure[AOI 2]{
	\includegraphics[height=0.35\linewidth]{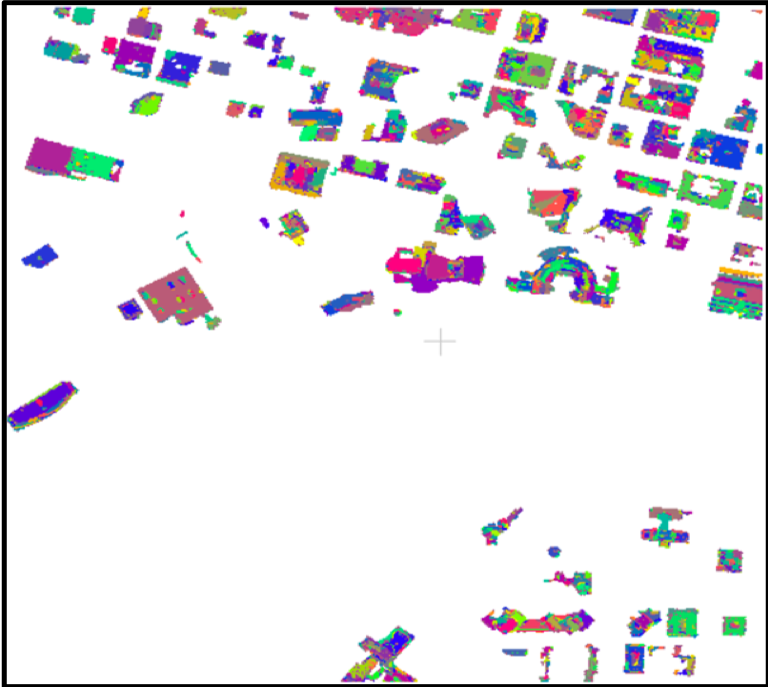}
	\label{fig:segAOI2}
	} \\
\subfigure[AOI 3]{
	\includegraphics[height=0.35\linewidth]{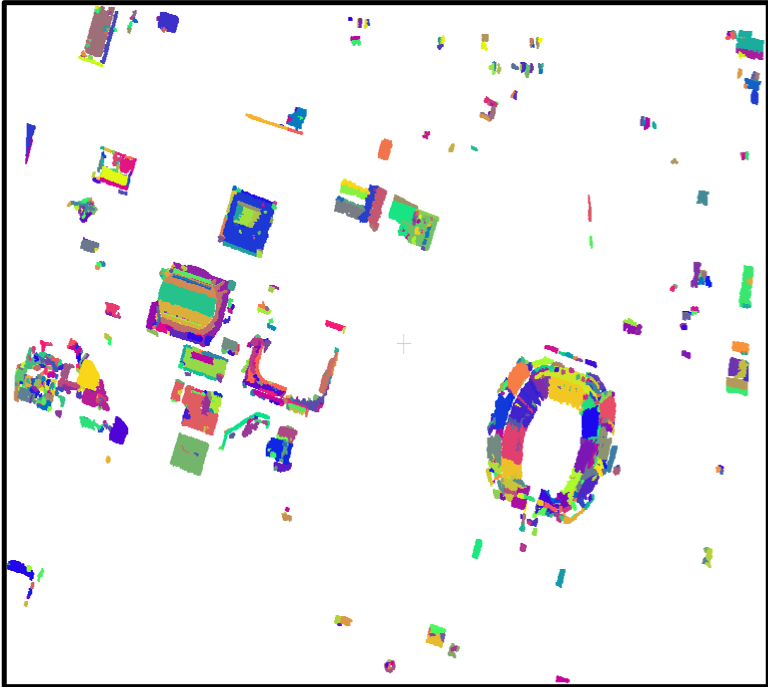}
	\label{fig:segAOI3}
	}
\hspace{2em}
\subfigure[AOI 4]{
	\includegraphics[height=0.35\linewidth]{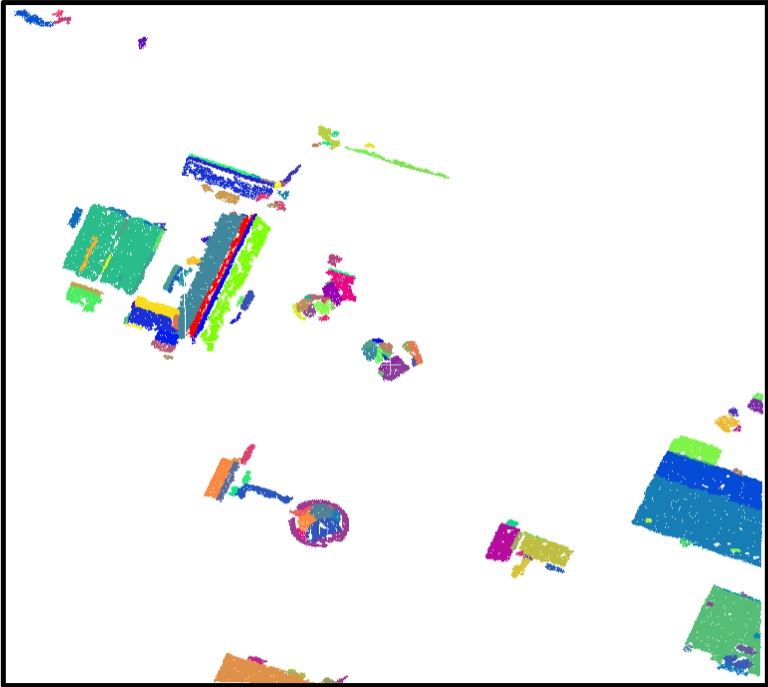}
	\label{fig:segAOI4}
	} \\
\caption{Segmented roof primitive planes from multi-cue hierarchical RANSAC (Different colors represent different primitives)}
\label{fig:segResults}
\end{center}
\end{figure}

\begin{figure}[ht]
\begin{center}
\subfigure[Google Image]{
	\includegraphics[height=0.22\linewidth]{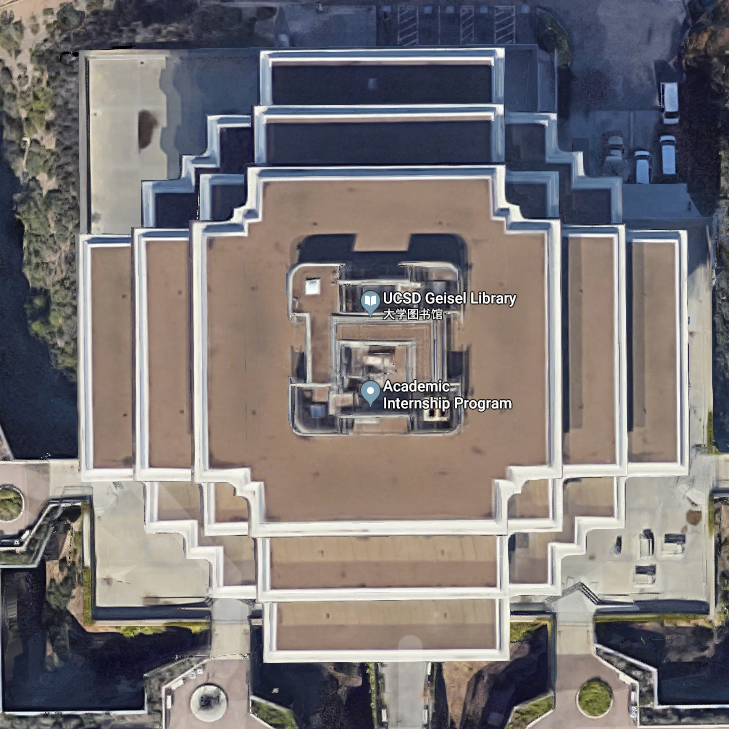}
	\label{fig:ucsdLibGoogle}
	}
\subfigure[Loose threshold]{
	\includegraphics[height=0.22\linewidth]{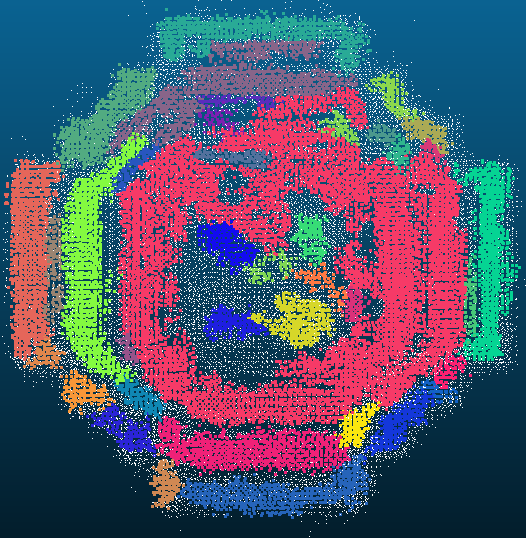}
	\label{fig:ucsdLibPCL1}
	}
\subfigure[Strict threshold]{
\includegraphics[height=0.22\linewidth]{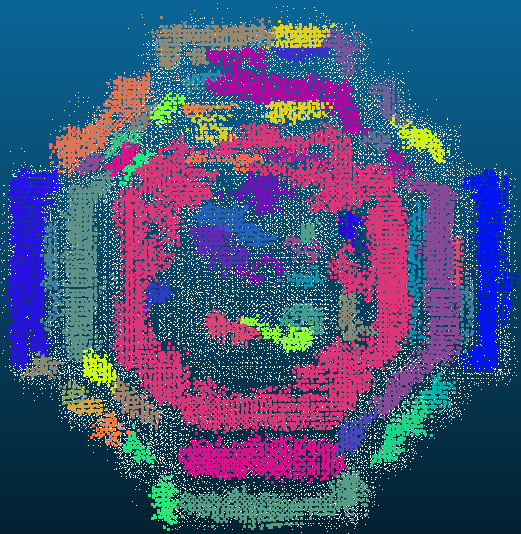}
\label{fig:ucsdLibPCL2}
	}
\subfigure[Ours]{
\includegraphics[height=0.22\linewidth]{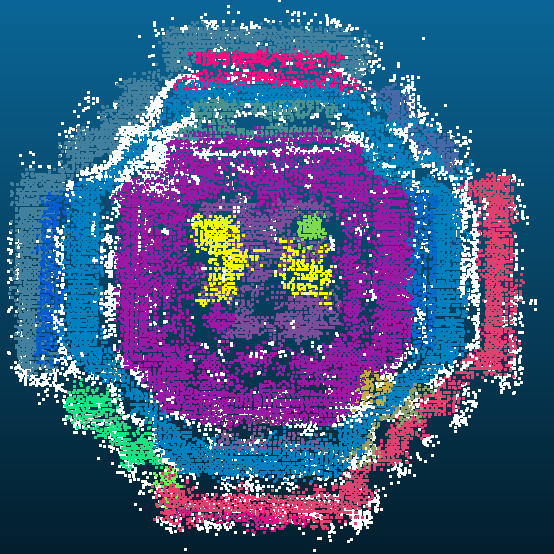}
\label{fig:ucsdLibOurs}
	} 
\caption{Roof segmentation comparison between the region growing method in the PCL Library and our multi-cue hierarchical RANSAC method (Different colors represent different primitives/planes)}
\label{fig:segCompare}
\end{center}
\end{figure}

\begin{figure}[ht]
\begin{center}
\subfigure[AOI 1]{
		\includegraphics[height=0.22\linewidth]{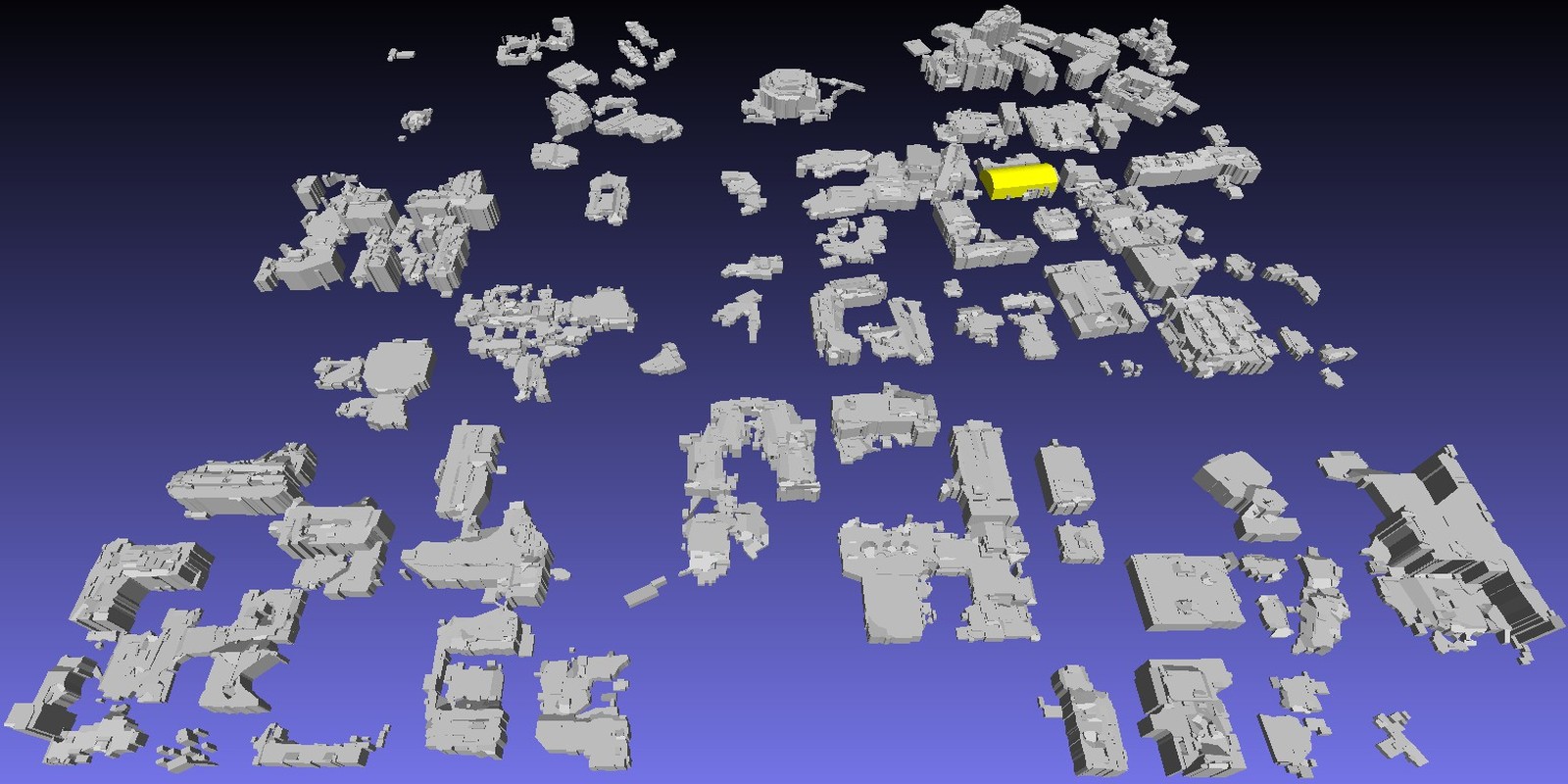}
	\label{fig:Model_AOI1}
	}
\subfigure[AOI 2]{
	\includegraphics[height=0.22\linewidth]{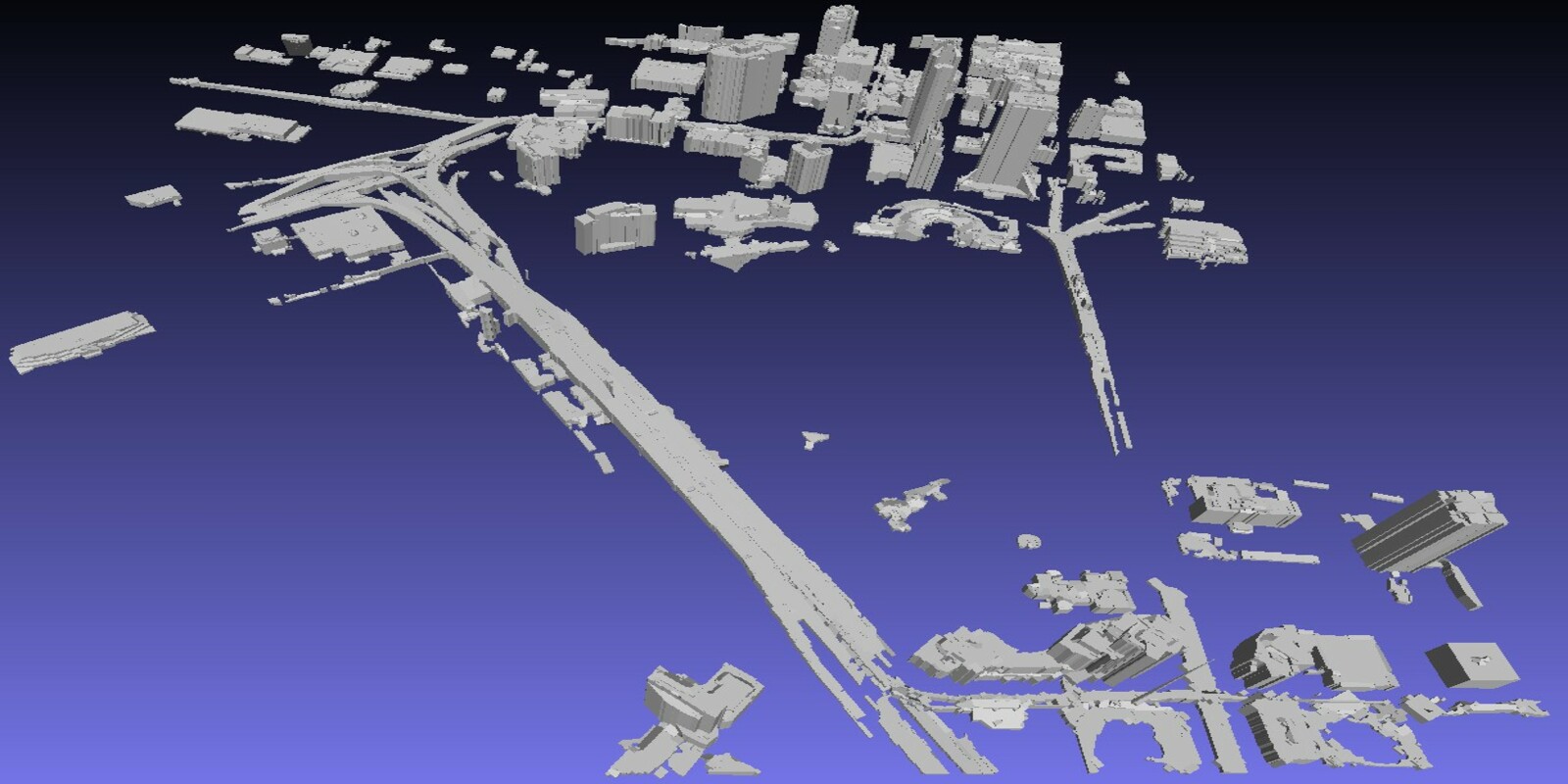}
	\label{fig:Model_AOI2}
	}
\subfigure[AOI 3]{
		\includegraphics[height=0.22\linewidth]{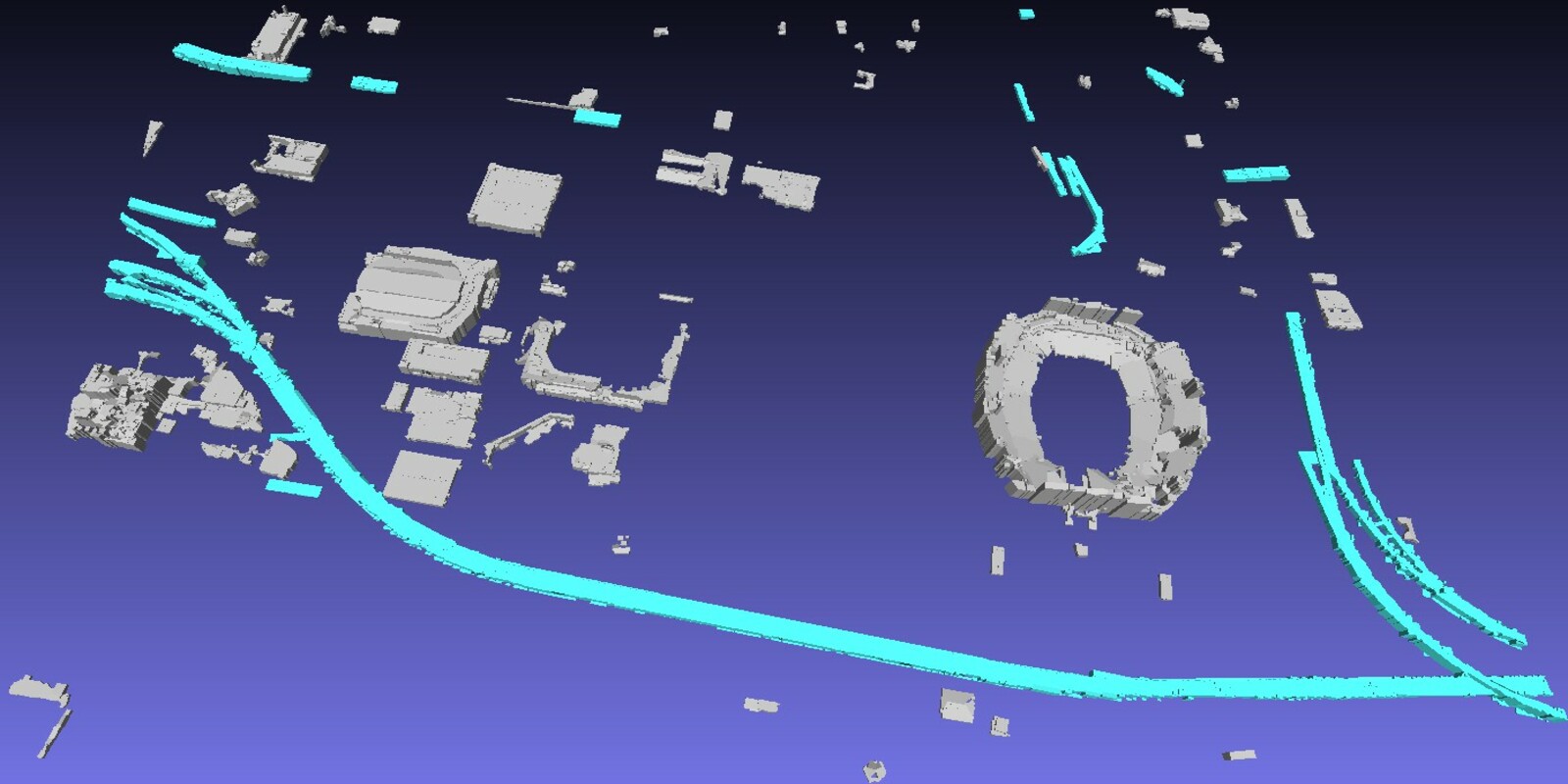}
	\label{fig:Model_AOI3}
	}
\subfigure[AOI 4]{
		\includegraphics[height=0.22\linewidth]{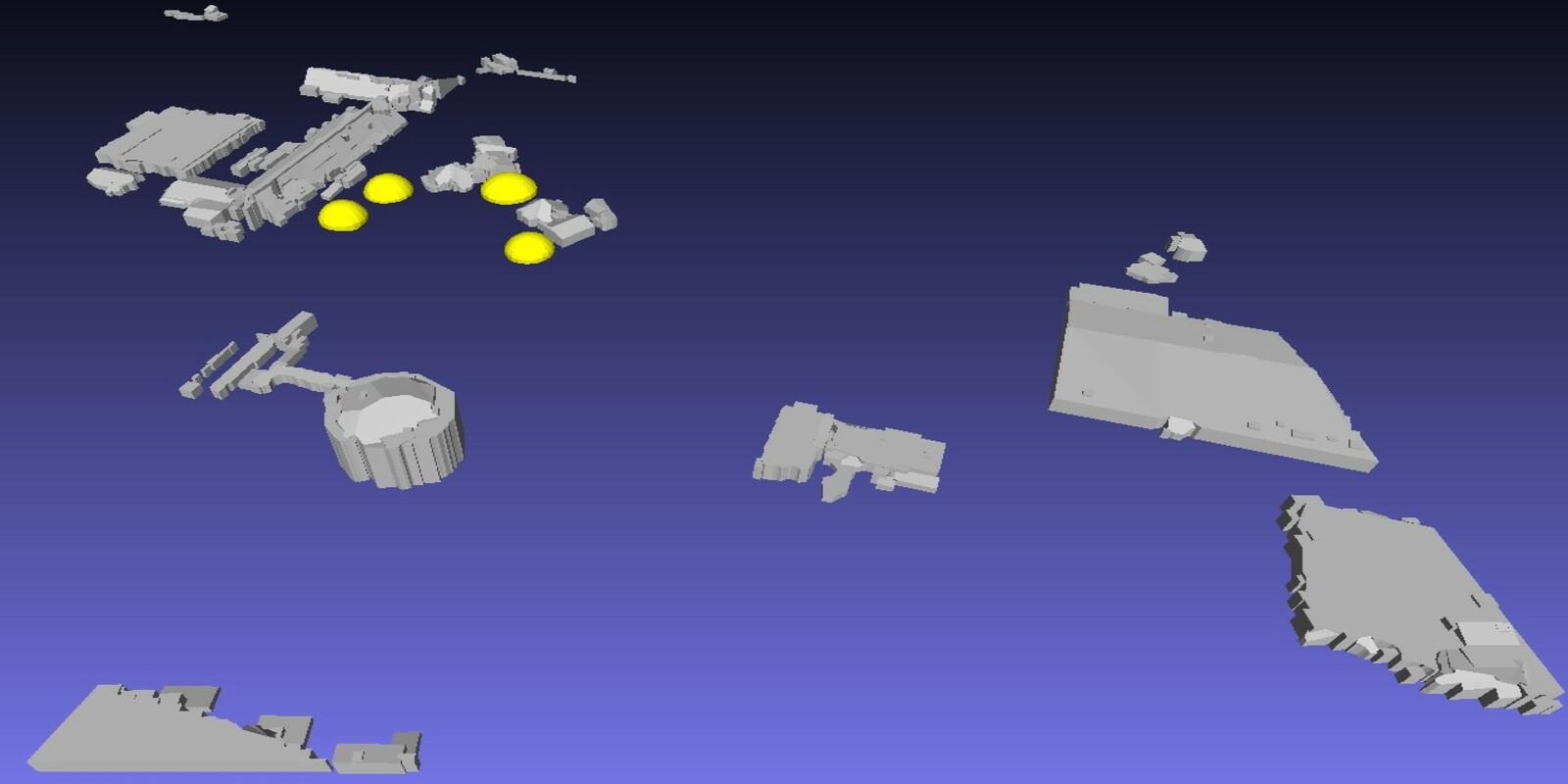}
	\label{fig:Model_AOI4}
	}
\caption{Building reconstruction results of the four AOIs}
\label{fig:ReconstructionResults}
\end{center}
\end{figure}

\subsubsection{Roof Primitive Segmentation and Fitting}
Fig.~\ref{fig:segResults} gives our overall segmentation results on four different AOIs using the 
proposed multi-cue hierarchical RANSAC. It is seen that we can generate fairly robust and detailed results even if the point cloud is very noisy. We compare the segmentation result of the proposed method to the results of the region growing based method in PCL with different threshold value~(Fig.~\ref{fig:segCompare}). 
The building in the image is the library of UCSD campus (in AOI1). 
As shown in the result, it is difficult to choose a proper threshold for the region growing methods in PCL library. 
Loose thresholds will result in under-segmentation whereas strict thresholds will produce many over-segmentation results. Because of the high data noise, it seems both over and under segmentation occur in the scene and no proper thresholds can satisfactorily balance both. Our multi-cue hierarchical RANSAC technique can be much more robust under such situation.

\subsubsection{Overall Reconstruction}

To evaluate the end-to-end performance of the proposed approach, we compare our reconstruction result with the ground-truth 2D building mask and the ground-truth DSM. Specifically, we render the reconstructed 3D building model back to a 2D binary building mask and a 3D DSM on top of the DTM and compare the ground truth of the 2D building mask and DSM. For both 2D and 3D, we apply 3 metrics, Completeness~(Comp., aka recall), Correctness~(Corr., aka precision) and Intersection over Union~(IoU) as defined in \citet{bosch2017metric}. 

Table.~\ref{tab:Final_precision} provides the overall reconstruction results of the AOI 1 and 2. The qualitative results are provided in Fig.~\ref{fig:ReconstructionResults}. Building models with complex roof shapes and various roof shapes under complex scenes are successfully created. This demonstrates the robustness of the proposed method.

\begin{table}[ht]
    \centering
    \begin{tabular}{c | c c c c c c c}
    \hline
    AOI  & comp 2D & corr 2D & IoU 2D & comp 3D & corr 3D & IoU 3D \\ \hline
    1  & 0.83 & 0.84 & 0.71 & 0.82 & 0.82 & 0.70\\  
    2  & 0.77 & 0.85 & 0.69 & 0.83 & 0.89 & 0.75\\ 
    \hline
    \end{tabular}
    \caption{Precision of the reconstructed models}
    \label{tab:Final_precision}
\end{table}

\begin{figure}
\begin{center}
\subfigure[Plane Models]{
	\includegraphics[height=0.20\linewidth]{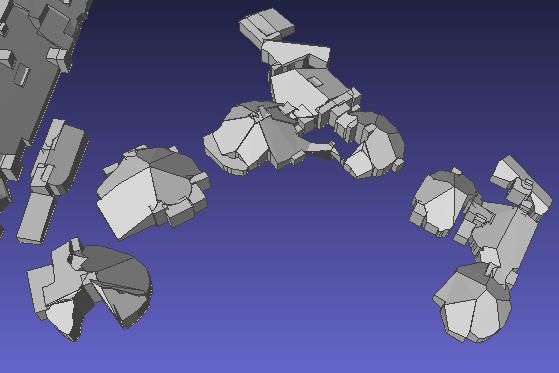}
	\label{fig:Highlight_Flat}
	} 
\subfigure[Our results]{
	\includegraphics[height=0.20\linewidth]{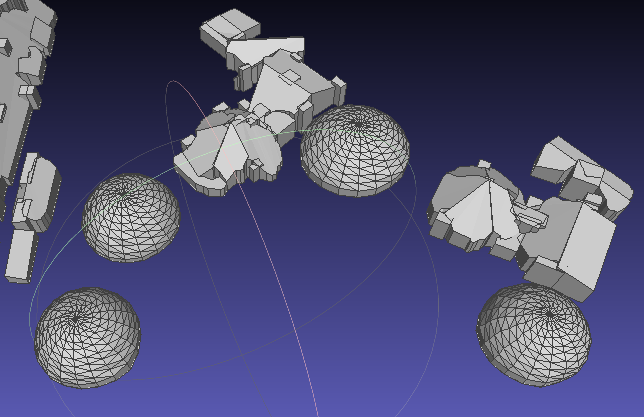}
	\label{fig:Highlight_Curved}
	} 
\subfigure[Google Map Models]{
	\includegraphics[height=0.20\linewidth]{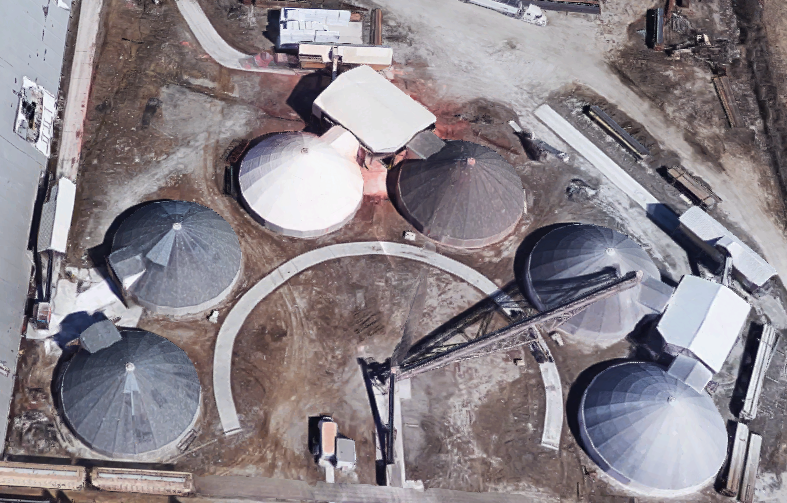}
	\label{fig:Highlight_Google}
    }
\vspace{-1.0em}
\caption{Zoomed-in view of the spherical shape roofs in Watco Omaha terminal in Omaha.}
\label{fig:Highlight}
\vspace{-1.0em}
\end{center}
\end{figure}

To show the capability of dealing with curved roofs with the proposed method, we highlight the spherical roofs in AOI 4 in Fig.~\ref{fig:Highlight}. We compare the 3D reconstruction result of our method~(Fig.~\ref{fig:Highlight_Curved}) with: the 3D reconstruction result of our method without cylindrical and spherical models~(Fig.~\ref{fig:Highlight_Flat}), and the 3D building model in Google Maps~(Fig.~\ref{fig:Highlight_Google}). The proposed method successfully captures 4 of the 6 sphere-shape roofs. Errors are due to the roof shape segmentation module. The model only using planar model produces a cracked result~(Fig.~\ref{fig:Highlight_Flat}).

\section{Conclusion} 
\label{sec:conclusion}

3D building reconstruction from point clouds created using satellite images is very appealing since the source data is relatively easy to acquire over large areas. However, due the high, orbital altitude of satellite observation, the 3D point clouds in urban areas generated from multi-view satellite  images suffer from a high level of structured noise and voids, both of which can be more severe than  in airborne data. These problems make the already difficult building reconstruction task more challenging, especially for large scale areas where diverse shapes of buildings may be present.  

To address these uncommon difficulties, we have designed an automated, robust, and end-to-end solution. Under the newly proposed deep learning guided 3D reconstruction framework,  we introduced recent developments in deep learning and extended traditional building reconstruction methods.  Roof shape segmentation was first carried out through PointNet network. A new data synthesis method was designed and applied effectively to directly learn from the point cloud. In the subsequent step, we further proposed a multi-cue hierarchical RANSAC to reliably extract roof primitives from the roof shape segmentation results. This allowed us to achieve a reliable and complete roof primitive segmentation. The final building reconstruction was completed through boundary regularization and roof topology.

Four complex urban areas with varying size from 0.34 to 2.04 square kilometers were used for evaluation. The proposed synthesized training method allowed the PointNet to achieved rather satisfactory results on roof shape segmentation that would otherwise require tedious human labeling. Moving least squares fitting and median filtering were necessary and could effectively suppress the intrinsic noise in the input point clouds. The outcome of the above steps provided a desired cleaned, void-free, and shape identified point cloud for the subsequent roof primitive segmentation.  The newly developed multi-cue RANSAC could take into account both the image colors and the surface normals, while the hierarchical RANSAC not only shortened the computation time but assured the robustness of roof primitive segmentation, leading to correct 3D reconstruction. It was demonstrated that an average of 83\% buildings can be assigned a correct shape. Quantitative evaluation with reference to airborne lidar data for two (0.96 and 2.04 sq km) of the larger areas reveals a 70-75\% overall IoU precision. It met the first expectation for an end-to-end pipeline for large scale complex city modeling in a fully automated environment. The implementation of the proposed algorithm is publicly available as an open-source software and can be deployed as an automatic service in Amazon Web Services. 

However, the final 3D reconstruction model is still inferior than that constructed from aerial image and LiDAR. Our future work will focuses on further improving the quality of the reconstructed models by integrating deep learning and model driven approaches.  

\section*{Acknowledgement}
This paper is supported by the Intelligence Advanced Research Projects Activity (IARPA) via Department of Interior/Interior Business Center (DOI/IBC) contract number D17PC00286. The U.S. Government is authorized to reproduce and distribute reprints for Governmental purposes not withstanding any copyright annotation thereon.

Disclaimer: The views and conclusions contained herein are those of the authors and should not be interpreted as necessarily representing the official policies or endorsements, either expressed or implied, of IARPA, DOI/IBC, or the U.S.Government.

\section*{References}

\bibliography{mybibfile}

\end{document}